\documentclass[conference]{IEEEtran}
\IEEEoverridecommandlockouts
\usepackage{cite}
\usepackage{amsmath,amssymb,amsfonts,upquote,textcomp,multirow,subcaption}
\usepackage[ruled,vlined]{algorithm2e}
\usepackage{algorithmic}
\usepackage{graphicx}
\usepackage{textcomp}
\usepackage{xcolor}
\def\BibTeX{{\rm B\kern-.05em{\sc i\kern-.025em b}\kern-.08em
    T\kern-.1667em\lower.7ex\hbox{E}\kern-.125emX}}
    
\usepackage{scalerel}
\usepackage{tikz}
\usetikzlibrary{svg.path}

\definecolor{orcidlogocol}{HTML}{A6CE39}
\tikzset{
  orcidlogo/.pic={
    \fill[orcidlogocol] svg{M256,128c0,70.7-57.3,128-128,128C57.3,256,0,198.7,0,128C0,57.3,57.3,0,128,0C198.7,0,256,57.3,256,128z};
    \fill[white] svg{M86.3,186.2H70.9V79.1h15.4v48.4V186.2z}
                 svg{M108.9,79.1h41.6c39.6,0,57,28.3,57,53.6c0,27.5-21.5,53.6-56.8,53.6h-41.8V79.1z M124.3,172.4h24.5c34.9,0,42.9-26.5,42.9-39.7c0-21.5-13.7-39.7-43.7-39.7h-23.7V172.4z}
                 svg{M88.7,56.8c0,5.5-4.5,10.1-10.1,10.1c-5.6,0-10.1-4.6-10.1-10.1c0-5.6,4.5-10.1,10.1-10.1C84.2,46.7,88.7,51.3,88.7,56.8z};
  }
}

\newcommand\orcidicon[1]{\href{https://orcid.org/#1}{\mbox{\scalerel*{
\begin{tikzpicture}[yscale=-1,transform shape]
\pic{orcidlogo};
\end{tikzpicture}
}{|}}}}

\newcommand\copyrighttext{%
  \footnotesize \textcopyright 2021 IEEE. Personal use of this material is permitted.
  Permission from IEEE must be obtained for all other uses, in any current or future
  media, including reprinting/republishing this material for advertising or promotional
  purposes, creating new collective works, for resale or redistribution to servers or
  lists, or reuse of any copyrighted component of this work in other works.}
\newcommand\copyrightnotice{%
\begin{tikzpicture}[remember picture,overlay]
\node[anchor=south,yshift=10pt] at (current page.south) {\fbox{\parbox{\dimexpr\textwidth-\fboxsep-\fboxrule\relax}{\copyrighttext}}};
\end{tikzpicture}%
}

\newcommand{\upquote}{\text{\textquotesingle}}

\PassOptionsToPackage{hyphens}{url}
\usepackage{hyperref} 

\begin{document}

\title{TinyOL: TinyML with Online-Learning on Microcontrollers}

\author{\IEEEauthorblockN{1\textsuperscript{st} Haoyu Ren\textsuperscript{\orcidicon{0000-0002-0241-6507}} \,}
\IEEEauthorblockA{\textit{Siemens AG} \\
\textit{Technical University of Munich}\\
Munich, Germany \\
haoyu.ren@siemens.com}
\and
\IEEEauthorblockN{2\textsuperscript{nd} Darko Anicic\textsuperscript{\orcidicon{0000-0002-0583-4376}}}
\IEEEauthorblockA{\textit{Siemens AG} \\
Munich, Germany \\
darko.anicic@siemens.com}
\and
\IEEEauthorblockN{3\textsuperscript{rd} Thomas A. Runkler\textsuperscript{\orcidicon{0000-0002-5465-198X}}}
\IEEEauthorblockA{\textit{Siemens AG} \\
\textit{Technical University of Munich}\\
Munich, Germany \\
thomas.runkler@siemens.com}
}

\maketitle
\copyrightnotice

\begin{abstract}
Tiny machine learning (TinyML) is a fast-growing research area committed to democratizing deep learning for all-pervasive microcontrollers (MCUs). Challenged by the constraints on power, memory, and computation, TinyML has achieved significant advancement in the last few years. However, the current TinyML solutions are based on batch/offline setting and support only the neural network's inference on MCUs. The neural network is first trained using a large amount of pre-collected data on a powerful machine and then flashed to MCUs. This results in a static model, hard to adapt to new data, and impossible to adjust for different scenarios, which impedes the flexibility of the Internet of Things (IoT). To address these problems, we propose a novel system called TinyOL (TinyML with Online-Learning), which enables incremental on-device training on streaming data. TinyOL is based on the concept of online learning and is suitable for constrained IoT devices. We experiment TinyOL under supervised and unsupervised setups using an autoencoder neural network. Finally, we report the performance of the proposed solution and show its effectiveness and feasibility.
\end{abstract}

\section{Introduction}
The Internet of Things (IoT) refers to the network of massive physical objects, \emph{things}, that are equipped with sensors, actuators, and tiny computers for sensing and communicating with each other. Microcontrollers (MCUs) are typically used as tiny computers as they are cheap and applicable in different domains. It is estimated that MCUs' worldwide shipment adds up to 30 billion per year\cite{b1}. The demand for MCUs is expected to grow steadily. 

Machine learning (ML) is another essential building block in IoT. Because of the wide deployment and the minimal energy consumption of IoT devices, people are increasingly paying attention to provide AI functionalities at the edge. The recent breakthroughs in Artificial Intelligence (AI) are driven by massive data, computation, and power consumption. For example, the carbon footprint for training the network "GPT-3" is approximately equal to traveling 700,000 kilometers by car\cite{b2}. However, MCUs are resource-constrained systems often powered by batteries. They are designed to live long with less than 0.1W power consumption and limited resources, e.g., 64MHz CPU frequency and 256KB RAM.

To fill the gap between MCUs and ML, especially deep learning, Tiny Machine Learning (TinyML\cite{b22}), as a rising AI field, is dedicated to offering solutions based on neural network (NN) at the edge. Remarkable results have been produced recently, such as voice and facial recognition\cite{b3}\cite{b4}. The key advantages of TinyML are related to:

\begin{itemize}
\item Privacy: The data is processed at the edge, whereas transmission of data to the cloud might violate privacy policies and is vulnerable to be intercepted. 
\item Latency: The whole process, which happens at the edge, is independent of external communication. Hence, edge devices can make decisions in real-time.
\item Energy efficiency: Powering NN is energy-intensive, yet transmission of the data to the cloud needs an order of magnitude more energy. 
\end{itemize}

Since most microcontrollers do not have an operating system, several bare-metal inference frameworks have been developed to support running NN on MCUs, while keeping computational overhead and memory footprint low. Some libraries are CMSIS-NN\cite{b23} from Arm and TensorFlow Lite for Microcontrollers\cite{b24} from Google. However, these libraries assume the model is trained in powerful machines or the cloud, and it is afterward uploaded to the edge device. The MCU only needs to perform inference.

This strategy treats the model as a static object. To learn from new data, the model must be retrained from scratch and re-uploaded to MCUs, making the deployment of TinyML in the industry environment a challenging task: 

\begin{itemize}
\item The embedded devices work in a distributed environment. Some devices are installed at remote and hardly accessible places. Considering the enormous amount of edge devices, updating the model in each of them is costly. 
\item Every machine is different, even if they are of the same type and from the same manufacturer. The model trained using the dataset from one machine might not work for another machine. 
\item The embedded devices have limited possibility to store field data. Transmission of field data back to the data center is expensive and subject to delay.  
\item The environment is constantly changing, the ML model's performance will drop if the input data distribution evolves, known as concept drift. The model's flexibility is constrained because MCUs can only execute the program's pre-defined routine. 
\end{itemize}

To tackle these challenges, we propose a novel system called TinyOL (TinyML with online-learning). TinyOL is implemented in C++, and it can be attached to an arbitrary existing network as an additional layer in MCUs. In addition to inference with the existing model, TinyOL can also learn from new data one by one and update the layer's weights in an online learning setting. Upon user request, TinyOL can train the layer or modify the layer structure to accommodate new data classes. With incremental learning, there is no need to store historical data for training. The model is always up to date and can thus deal with the concept drift.

We prove our approach by providing two use cases using autoencoder NN and running it on an Arduino Nano 33 BLE Sense board\cite{b25}, which is embedded with Cortex™-M4 CPU. Specifically, we attach TinyOL to an existing autoencoder in the MCU,  enable the onboard post-training, and teach the model to classify new data patterns. The experimental results show that the TinyOL system can help NN adapt to the new environment by learning from field data and improves the pre-trained model's performance incrementally. Furthermore, TinyOL can provide flexibility to the bare metals on the fly, e.g., it gives unsupervised autoencoder the classification ability. To the best of our knowledge, this is the first paper that brings incremental online learning in the world of MCUs, and it proves the feasibility of on-device post-training and remodeling of pre-trained NN. 

The structure of the remaining work is as follows. The next section, Section \ref{section:related_work}, discusses the literature reviews regarding TinyML, online learning, and on-device training. This is followed by Section \ref{section:approach}, which includes our proposed solution and its methodologies.  In Section \ref{section:evaluation}, the experimental results are presented and analyzed.  Lastly, Section \ref{section:conclude} provides the conclusion and potential areas for future research.

\section{Related Work}
\label{section:related_work}

The main research fields in ML on tiny devices can be described into three perspectives: 
\begin{itemize}
\item Deep learning algorithm at the edge: Designing better algorithms that can utilize resources more efficiently without compromising the performance, including compressing pre-trained NN\cite{b5}\cite{b6} and designing or searching better architectures\cite{b7}\cite{b8}.
\item TinyML hardware: Many ultra-low-power hardwares have been proposed recently, which can be efficiently powered by a battery for an extended period. \cite{b9}\cite{b10}.
\item TinyML application: Optimizing the deployment of TinyML on MCUs in the real environment\cite{b11}.
\end{itemize}

Apart from the research activities, several
light-weighted libraries have been developed to support NN's inference optimization at the edge, including open-source frameworks: TFLite Micro from Google, CMSIS-NN from Arm, TVM\cite{b26} from Apache, and the framework that only supports proprietary devices: X-CUBE-AI\cite{b27} from STM. However, all of these libraries assume the NN is first trained in a powerful machine and later be flashed to MCU as a C array without supporting on-device training.

For online machine learning, less attention is drawn than traditional batch/offline learning because researchers tend to think all the training data are available at the training phase. In \cite{b12}, a broad convolutional neural network with incremental learning capacity is proposed to learn from newly collected abnormal samples. A few works \cite{b13}\cite{b14} discuss how to against catastrophic forgetting problems of incremental online learning in NN. The river\cite{b15} is a python library with the ambition to run traditional machine learning on streaming data.   

The NN training at the edge is much less common than inference because of limited resources and reduced numerical precision.  According to \cite{b16}, good performance can only be guaranteed when training and inference run on the same dataset. The accuracy of the model will decrease over time. Therefore, the author suggests retraining the network on the device regularly, in their case - Raspberry Pi. In \cite{b17}, a shallow NN is implemented at the edge for inference, powered by an NVIDIA Jetson. The edge node is connected with a deep NN that resides in the server. To save energy, deep NN is only used to transfer knowledge to shallow NN when a significant drop of performance at the edge is detected. Similarly, a near real-time training concept is proposed in \cite{b18} for computer vision tasks on NVIDIA Jetson.  Whenever a new object class is detected, a new NN model is initialized using the old model's weights, retrained at the fog, and be ported back to the edge.  In another work \cite{b19}, the checkpointing strategies are adopted to save training consumption at the edge, whereas in \cite{b20}, incremental learning is applied to support the training of the k-nearest-neighbor model on Raspberry Pi. In \cite{b21}, expert feedbacks are provided as labels to an unsupervised NN on the fly, thus turning unsupervised learning into semi-supervised learning.

To sum up, although AI has been brought into play in the MCUs world and many great works have been produced, there is still a gap between MCU and ML, namely, training the NN on bare metals. All the aforementioned TinyML solutions assume that only inference is needed at the edge. Some works introduce training NN at the edge, but the "edge" referred in their works has an order of magnitude more available resources than constrained MCUs. Hence, their methods are not applicable for tiny edge devices.

\section{Approach}
\label{section:approach}
Creating a general NN model for all is not a viable solution for MCUs. Different users have different patterns, and the model needs to be updated with field data on the fly. However, training model consumes a lot of resources, which hinders the promotion and practicality of TinyML on resource-constrained devices. 

\begin{figure}[htbp]
\centerline{\includegraphics[scale=0.65]{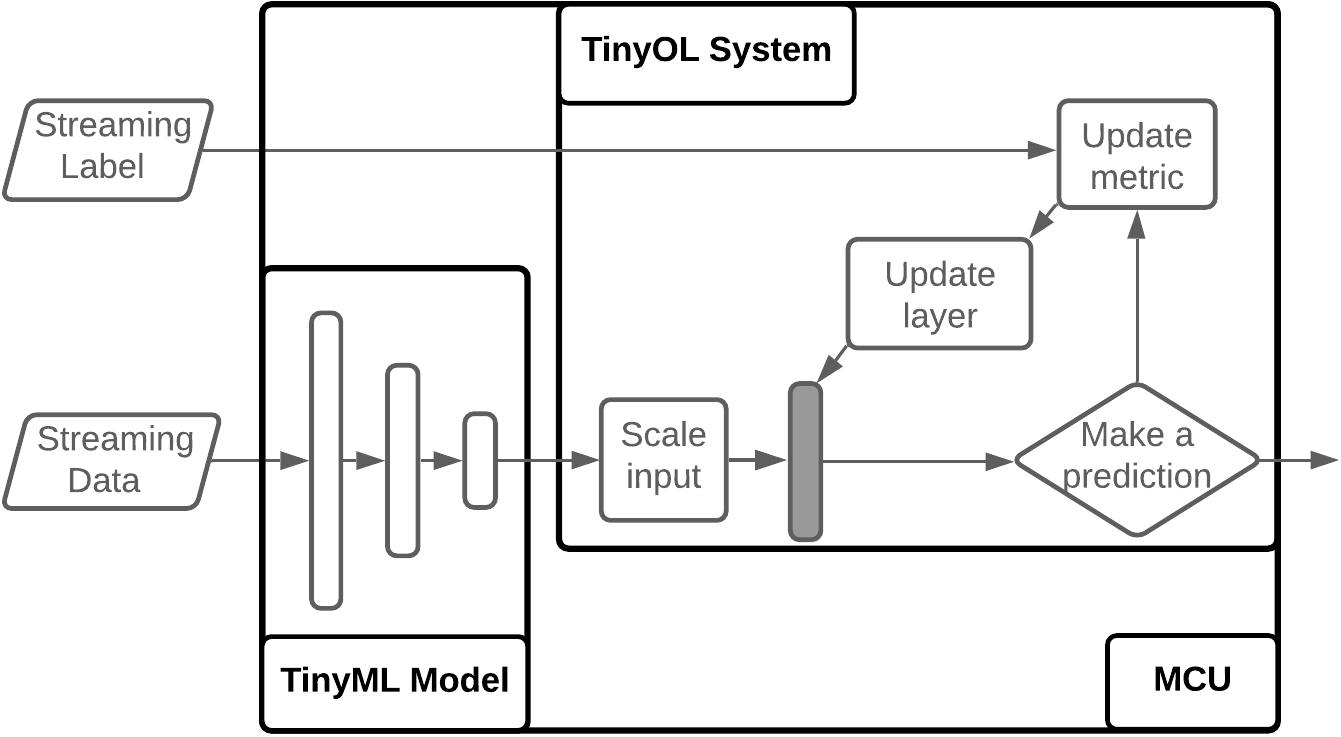}}
\caption{The building blocks of TinyOL on MCUs.}
\label{f1}
\end{figure}

The proposed system can be attached to an existing NN in MCUs as a new layer or replace a specific layer in the NN, as shown in Fig.~\ref{f1}. When the data stream comes, TinyOL can inference every upcoming sample and subsequently update its weights by exploiting online learning's advantages. If objects of a new class are detected, TinyOL can modify the output layer's structure to accommodate the latest information. Thus, many constraints on MCUs, i.e., data storage, computation overhead, can be untiled.

\subsection{The TinyOL System}
Running inference of NN on MCUs is equivalent to process streaming data one after another. The NN can process the next input sample after the finish of the last inference.  TinyOL utilizes this characteristic and further extend the existing NN with post-training ability. 

The core component of TinyOL is the additional layer marked in black in Fig.~\ref{f1}. This extra layer consists of several neurons, which can be customized, initialized, and updated on the fly. After applying the system on top of other existing networks, the additional layer becomes a part of the NN as a new output layer. Because the existing NN are uploaded as a C array to MCU's Flash, they are treated as a frozen graph and cannot be modified afterward. However, the neurons in the additional layer can be trained since TinyOL runs in the RAM.  This concept is similar to transfer learning, where a part of a pre-trained model is fixed, and fine-tune happens at the last few layers. 

With the online learning architecture, the additional layer can learn with streaming data. The pseudo-code of the TinyOL pipeline is summarized in Algorithm~\ref{a1}. At each time, a new sample data first flows through the existing NN and subsequently fed into TinyOL. Depending on the tasks, the accumulated mean and variance will be updated, and the input can be standardized. Next, the system will run an inference. If a corresponding label is available, the evaluation metrics and the weights in the additional layer will be adapted using online gradient descent algorithms, e.g., stochastic gradient descent (SGD). Thus the training and prediction steps are inter-leaved. Once the neurons are updated, the sample pairs can be discarded effectively. In other words, at a time, only one data pairs of the stream live in the memory, and there is no need to store the historical data. Compared to the batch/offline training setting, TinyOL can be trained with minimal resource usage, making on-device training on massive streaming data possible. 


\begin{algorithm}[htbp]
  \caption{TinyOL Workflow}
  \begin{algorithmic}[1]
    \STATE Initialize TinyML model and TinyOL system;
    \FOR{x in StreamingData} 
        \STATE x\upquote \ = TinyML.Process(x);
        \STATE TinyOL.UpdateRunningMeanAndVariance(x\upquote);
        \STATE x\upquote \upquote \ = TinyOL.ScaleInput(x\upquote);
        \STATE y\upquote \ = TinyOL.Predict(x\upquote \upquote);
        \IF{y is available}
          \STATE TinyOL.UpdateMetrics(y\upquote, y); 
          \STATE TinyOL.UpdateWeights(y\upquote, y);
        \ENDIF
    \ENDFOR
  \end{algorithmic}
  \label{a1}
\end{algorithm}

With this design, models can be adapted to a specific field because the layer in TinyOL can be trained in real-time upon the arrival of streaming field data. The consequence is that the model is robust against the concept drift, which implies that the field data's statistical properties might vary over time. The model's performance will drop significantly without post-training because the model cannot foresee these changes in the training phase.





\subsection{Concept Design}

\begin{figure}[htbp]
\centerline{\includegraphics[scale=1]{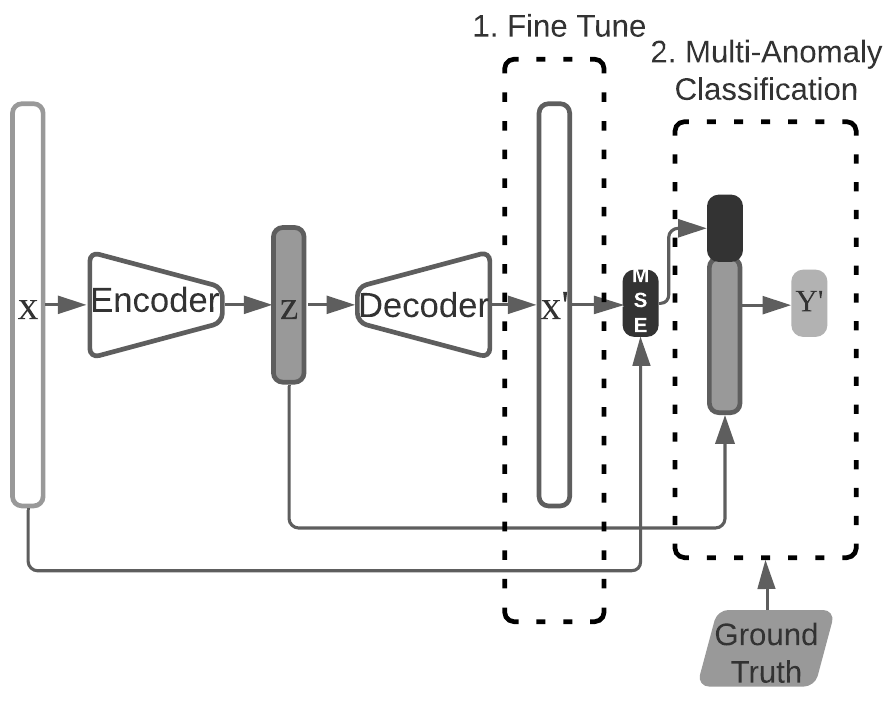}}
\caption{The TinyOL system and the autoencoder synergy, where the two dashed blocks correspond to two experimental designs.}
\label{f3}
\end{figure}

To assess the TinyOL concept, we design two use cases under unsupervised and supervised settings to illustrate how the proposed method can alleviate industry problems. The system is incorporated into an existing NN - anomaly detection autoencoder, shown in Fig.~\ref{f3}. 

An autoencoder is an unsupervised neural network trained to reconstruct the input data $X$ after compressing it into an intermediate embedding $Z$. The network consists of two parts: an encoder $p_{encoder}(Z|X)$ that learns to map high dimensional data into low dimension representation, and a decoder $p_{decoder}(X\upquote|Z)$ that reconstructs the input from the embedding. The loss function, typically the mean square error (MSE), is used to assess the differences between input and the reconstructed output:

\begin{equation}
    MSE = \frac{1}{n}\sum\limits_{i=1}^n(X_i-X\upquote_i)^2,
\end{equation}

where $n$ is the number of the input and output dimensions.

In the industry, we can apply autoencoder to detect anomalies by training only on the machine's normal data. The network should learn to work well with normal samples, which means low reconstruction error. After the deployment, if an observation deviates too much from the regular training data, the reconstruction error will be much higher. By setting a threshold for the reconstruction error, we can separate the abnormal data from the normal data. 

In the following, we explain the design of two use cases. As shown in Fig.~\ref{f3}, the first use case aims to demonstrate TinyOL's ability to fine-tune the existing NN on MCUs. The second examines the feasibility of turning an unsupervised autoencoder into a supervised anomaly classification model by modifying the layer structure at runtime.

\subsubsection{Use case – fine-tune}

 In the real application, we face the problem that the field data might not be the same as the training data, as each field has its own signature. The concept drift will disrupt the performance of the underlying system. 
 
Therefore, it is essential to democratizing MCUs to post train their NN on the fly. In the first use case, the existing autoencoder's final layer is replaced by TinyOL, which takes responsibility for post-training the last layer in online mode. 

Assume activation function of the last layer is sigmoid function, the output of decoder is $A$, and the weights between the decoder and last layer are $W$. The forward propagation of one connection in the last layer is written as:

\begin{equation}
    x\upquote = \sigma(a_1 \cdot w_1 + … + a_n \cdot w_n),
\end{equation}

\begin{equation}
    \sigma(a) = \frac{1}{1+e^{-a}}.
\end{equation}
          
Take cross entropy loss function as an example:

\begin{equation}
     L(x, x\upquote) = -(x \cdot log(x\upquote) + (1-x) \cdot log(1-x\upquote)).
\end{equation}
             
Subsequently, we apply chain rule to update each weight in the connection: 

\begin{equation}
    w_i := w_i - \alpha \cdot (x\upquote – x)\cdot \sigma(x\upquote)\cdot (1-\sigma(x\upquote)) \cdot a_i,
\end{equation}

where $\alpha$ is the learning rate.
             
Note that because the system is running in online mode, only one sample will be processed at a time, which is different from the offline setting, where all the samples' losses are accumulated.

\subsubsection{Use case – multi-anomaly classification}

Often, an engineer wants to monitor the machines' anomalies and interpret the irregularities' details. However, the categorical anomaly information is not embedded in the machine's program by manufacturers as they don't know which abnormality will occur and its character before deployment. By leveraging TinyOL's flexibility, we enable the classification of different anomaly patterns on the go. 

In the second use case, an under-complete autoencoder strategy is applied, where the output from the encoder is extracted together with the reconstruction error as the classification features. The dimension of the embedding $Z$ is usually much smaller than the input dimension, designed to capture the most prominent features and reduce data dimension. Eventually, the encoder's output and the reconstruction error are scaled by running variance and mean, and fed into the TinyOL system as input features. 

Assume different anomaly patterns need to be classified, the TinyOL system will initialize a multi-class classification layer. It maintains a weights vector for each class and will initialize extra neurons when a new class appears. Suppose we choose a softmax regression layer for the task. We denote the input feature as $X$ and the corresponding weights in the multi-class layer as $W$. The forward-propagation of one class is:

\begin{equation}
    y\upquote = \frac{e^{w^T_{y_j}x}}{{\sum_{j=1}^{k}} e^{w^T_{y_j}x}}.
\end{equation}
 
Again, the cross-entropy loss function for one data sample is chosen as an example:

\begin{equation}
    L(Y, Y\upquote) = -\sum_{j=1}^{k}(1\{y^i=k\} \cdot log(\frac{e^{w^T_{y_k}x}}{\sum_{j=1}^{k}e^{w^T_{y_j}x}})),
\end{equation}

where $k$ is the number of the current anomaly classes.

If a ground truth label $y$ is provided for this sample, for every parameter in one class, then we update the weight incrementally:

\begin{equation}
    w_i := w_i - \alpha \cdot x^i \cdot (\frac{e^{w^T_{y_k}x}}{\sum_{j=1}^{k}e^{w^T_{y_j}x}} - 1\{y=k\}),
\end{equation}

where $\alpha$ is the learning rate.

\begin{figure}[htbp]
\centerline{\includegraphics[scale=0.07]{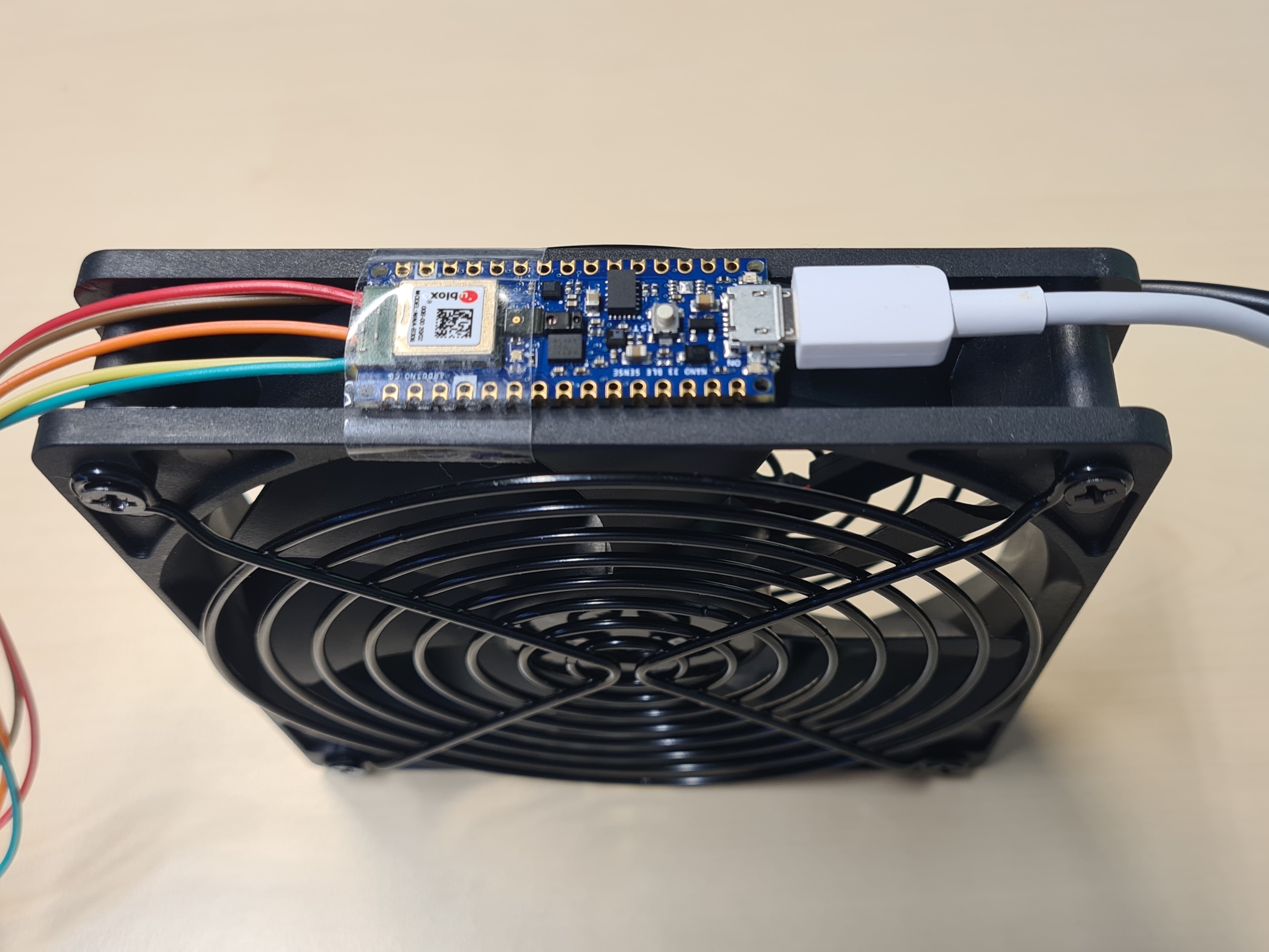}}
\caption{The hardware setting: the Arduino board and the USB fan.}
\label{pic1}
\end{figure}
             
At each time, only one data sample is used to update the weights. If a new class label is provided, TinyOL will initialize additional connections in the system automatically.

\subsection{Experimental Design and th Data}

\begin{figure}[htbp]
\centerline{\includegraphics[scale=0.42]{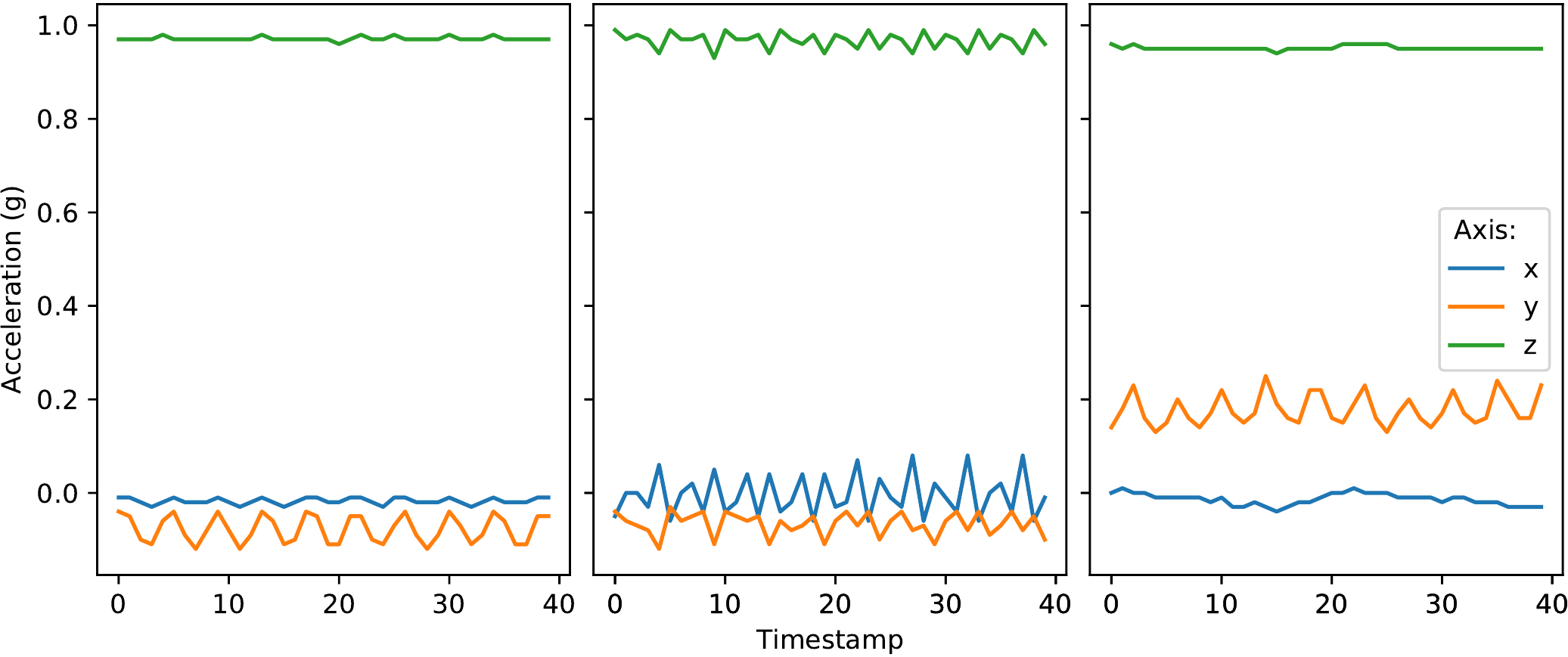}}
\caption{The 3-axis vibration pattern. Left: normal. Middle: stuck. Right: tilted.}
\label{f10}
\end{figure}

To evaluate the concepts mentioned above, we conduct experiments using an Arduino Nano 33 BLE board\cite{b25} and a USB fan, as shown in Fig.~\ref{pic1}.  The board is featured with a Cortex™-M4 CPU running at 64 MHz, 256KB SRAM, whose specification lies in the typical range of MCUs.  Moreover, it comes with several embedded sensors, in which the 3-axis accelerometer sensor is used in the experiment. The fan is used to simulate an industrial rotating machine. Specifically, we use the board to read three different vibration patterns from the USB fan: regular operation, the fan is stuck, and the fan is tilted. Fig.~\ref{f10} plots the 3-axis vibration data of each class for 40 timestamps, corresponding to about 0.25 seconds using the accelerometer with a 119Hz sampling rate. The board will integrate TinyOL into the pre-trained autoencoder to run inference, post-training, and classifying the fan's operation mode at runtime.

The experimental design is arranged as follows:

\begin{enumerate}
\item Collect vibration data of USB fan from three categories: normal operation, stuck, and tilted. For each category, five minutes of data are recorded. The normal data is used to train the autoencoder, whereas the abnormal data is used for evaluation purposes. We use PCA to reduce the 3-axis dimension data to one dimension. The autoencoder architecture is chosen with an input size (40, 1) after several attempts by trial and error. The embedding layer has four dimensions, which balances well between the latency and the performance of the model.


\item Train an autoencoder on the normal data, post-optimize it into a C array using TFLite Micro, and upload the NN with the TinyOL system to the board.
\item Validate the first concept – fine-tune the model in the online mode.
\item Validate the second concept – train the model to classify different patterns of the fan on the fly.
\end{enumerate}

\begin{figure}[htbp]
  \centering%
    \begin{subfigure}{0.81\columnwidth}
    \includegraphics[width=1\linewidth]{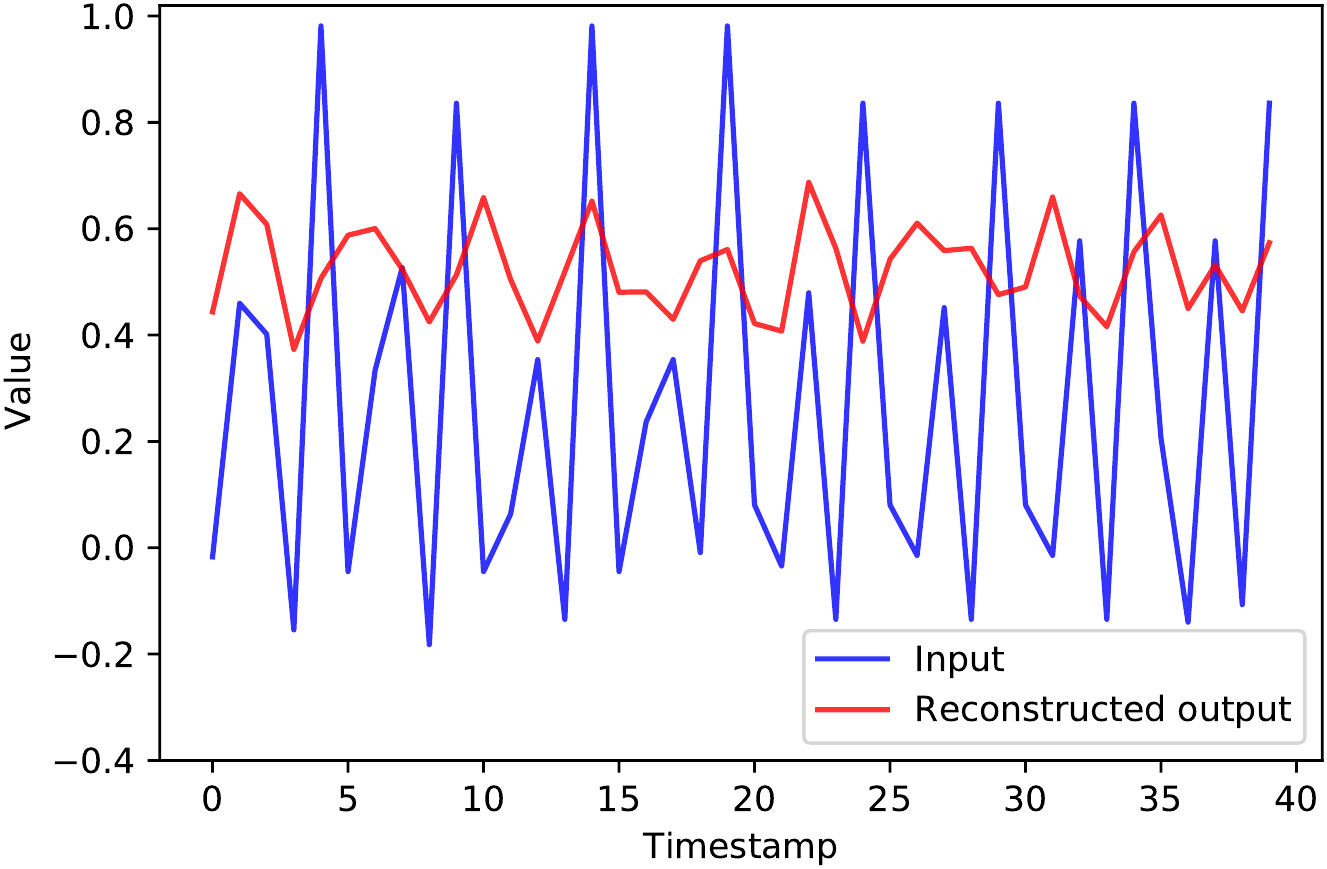} 
    \caption{Abnormal sample.} 
    \end{subfigure} 
    \begin{subfigure}{0.81\columnwidth}
    \includegraphics[width=1\linewidth]{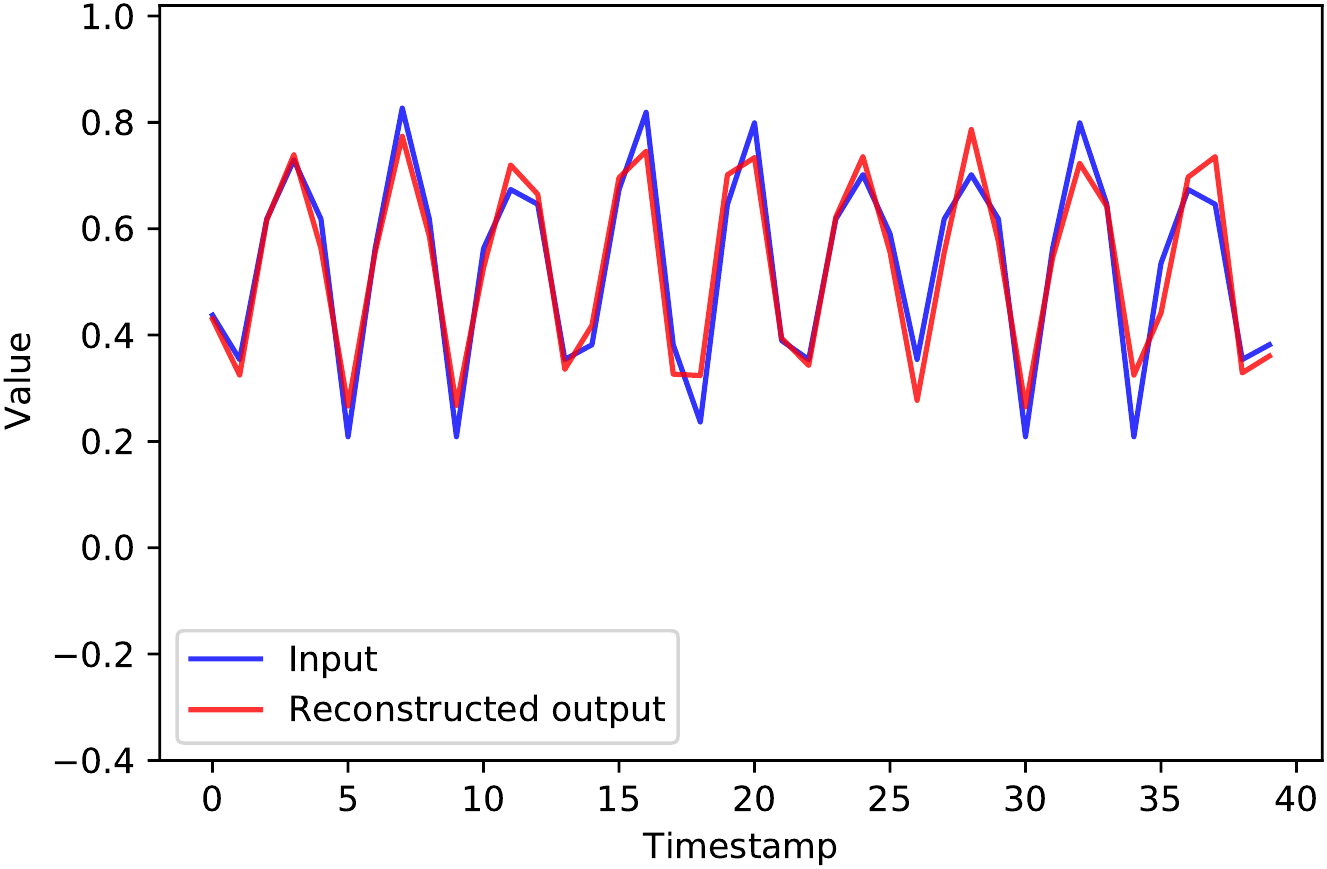} 
    \caption{Normal sample.} 
    \end{subfigure} 
    \caption{The input and the corresponding reconstructed output from the trained autoencoder.} 
    \label{f4}
\end{figure}

\section{Evaluation}
\label{section:evaluation}

This section describes the experimental results in the following steps: firstly, we present the autoencoder's performance trained on the laptop. Secondly, the model is tested on the board, where we illustrate the concept drift problem. By overcoming the concept drift, the effectiveness of TinyOL is evaluated and emphasized. Finally, we highlight the system's online classification feature with a statistical comparison to the batch/offline training.

\subsection{Autoencoder}

\begin{figure}[htbp]
\centerline{\includegraphics[scale=0.65]{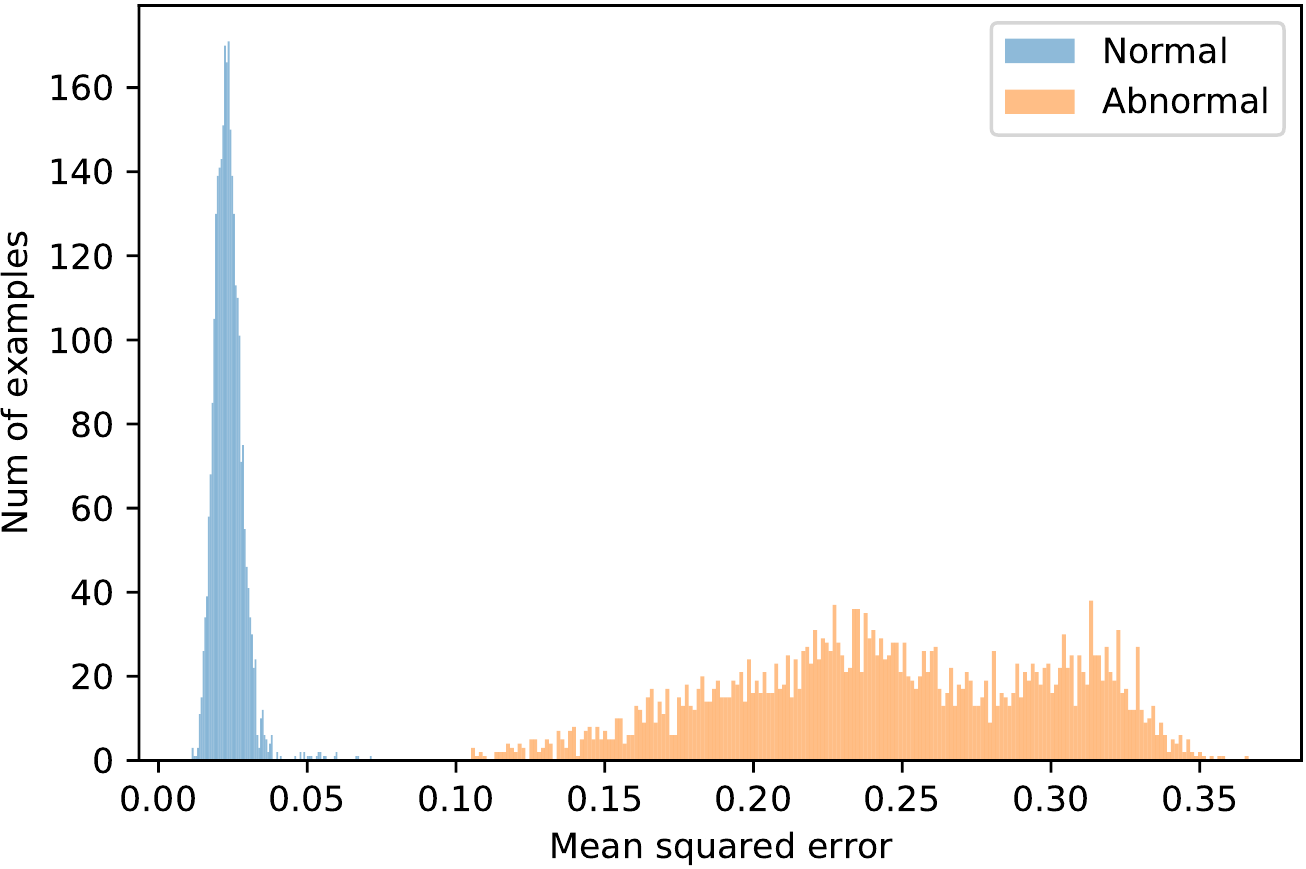}}
\caption{The distribution of the MSE from 3000 normal and 3000 abnormal samples.}
\label{f6}
\end{figure}

We train the autoencoder on normal data samples to minimize the reconstruction error. The hypothesis is that abnormal data samples (in our experiment: tilted and struck) will have a higher reconstruction error than normal data samples. Fig.~\ref{f4} plots input (after PCA) and the autoencoder's reconstructed output, where the upper figure describes an anomalous example, and the lower plot is for a normal sample. Fig.~\ref{f6} shows a histogram representing the different distributions of reconstruction error from normal and abnormal data. One can observe that the mean squared error of anomalous samples is much higher than the normal data. We can set a fixed value as a threshold in the industry application, where an anomaly is classified when the error surpasses that threshold.

\subsection{Fine-Tune}

After the autoencoder is proved to work technically, we flash the model to the board for testing in real scenarios. 

\begin{figure}[htbp]
  \centering%
    \begin{subfigure}{0.95\columnwidth}
    \includegraphics[width=1\linewidth]{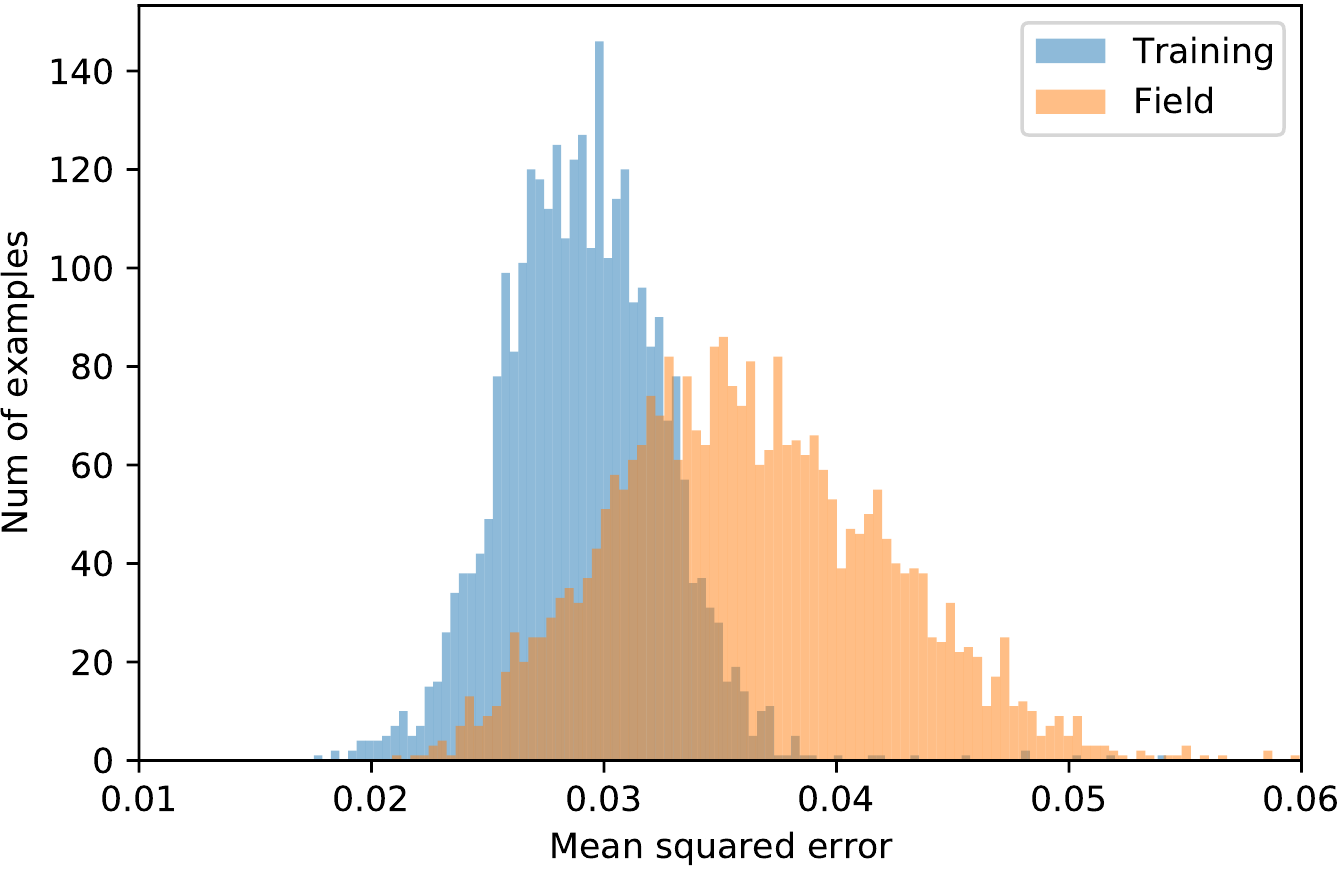} 
    \caption{Before the fine tuning.} 
    \end{subfigure} 
    \begin{subfigure}{0.95\columnwidth}
    \includegraphics[width=1\linewidth]{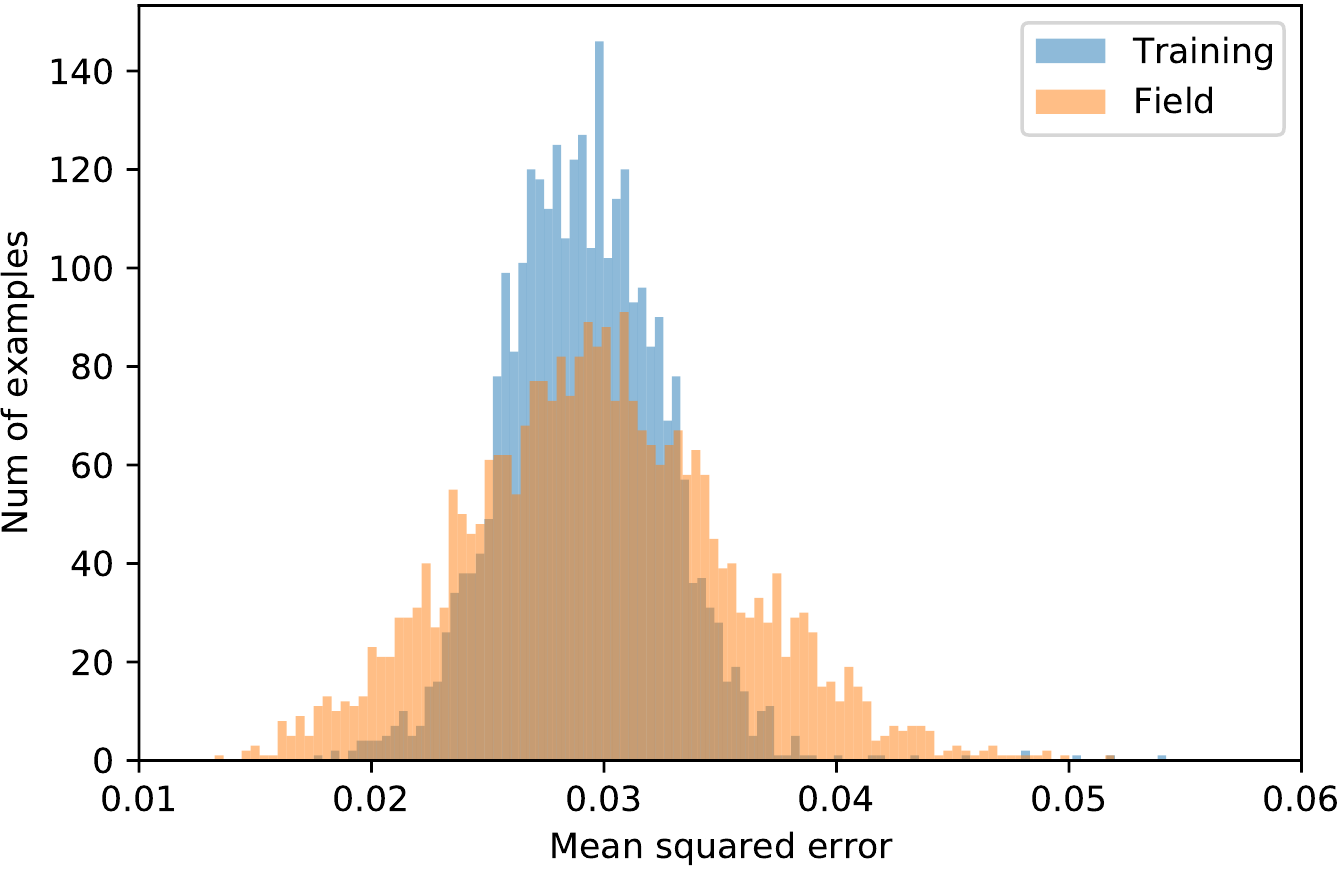} 
    \caption{After the fine tuning.} 
    \end{subfigure} 
    \caption{The distribution of 3000 normal samples' MSE collected in the field (orange) compared to 3000 normal samples' MSE collected at training phase (blue).}
    \label{f7}
\end{figure}

The board is put back to the fan, to the position where we collected training data. However, the upper histogram in Fig.~\ref{f7} shows different distributions of the reconstruction error between the pre-collected data from the training phase (in blue) and the real field data after we place the board back to the fan (in orange).   Although we try to put the board back to the position where the training data are collected as precisely as possible, it is unavoidable that minor deviations exist. The autoencoder is very sensitive to the data measured from different positions, and therefore, delivers a different distribution of the MSE. If we use a fixed value as a threshold for anomaly detection, the algorithm fails because too many false alarms will be generated.  In other words, we cannot guarantee that the training data and the inference data have the same statistical property in the real industry setting. 

Different machines have different patterns, and models need to be updated using data specific to the appliances. Customizing and training NN for every user from scratch is time-consuming. Thus, we evaluate the TinyOL system's post-training feature by replacing the last layer of the autoencoder with the additional layer in the system. We enable the on-board online post-training for 2000 iterations via Bluetooth. Each time,  a streaming sample is used to fine-tune the weights in TinyOL. The model is evaluated after that, as shown in the lower histogram in Fig.~\ref{f7}. We observe that, after fine-tuning, the reconstruction error in the field has a similar distribution as the data collected at the training phase.  It demonstrates the NN should be post trained locally at the edge to fit into different working conditions.

We compare the time consumption for one iteration between inference and online learning on 3000 samples in Table.~\ref{t0}. In the online learning setting, the MCU first runs an inference and subsequently updates the weights in TinyOL. In this experiment, the ground truth is the same as the input. The result indicates that on-device incremental training consumes a little bit more time. However, the post-training is not enabled for lifelong. Once a satisfied accuracy is achieved, we can turn it off. Without many compromises of computing delay, we can guarantee good performance for the individual model.

\begin{table}[htbp]
\caption{The comparison of time consumption for each iteration on 3000 samples (µs)}
\begin{center}
\begin{tabular}{|c|c|c|c|c|}
\hline
\multirow{2}{*}{\textbf{Mode}} & \multicolumn{4}{c|}{\textbf{Metrics}} \\ \cline{2-5} 
 & \textit{\textbf{Average}} & \textit{\textbf{Medien}} & \textit{\textbf{Minimum}} & \textit{\textbf{Maximum}} \\ \hline
Inference                      & 1748    & 1648    & 1403    & 1991    \\ \hline
Online Learning                & 1921    & 1800    & 1404    & 3357    \\ \hline
\end{tabular}
\end{center}
\label{t0}
\end{table}

\subsection{Multi-Anomaly Classification}
Next, we explain the second experiment for TinyOL: use the pre-trained autoencoder to classify the fan modes at runtime. Specifically, we use the encoder's output and the construction error as input features to classify the normal, stuck, and tilted fan status incrementally. 

We put the fan into different modes and activate the attached board to learn the class pattern via Bluetooth. We train the model for several iterations on the streaming vibration data from one class, then we switch the fan to a different mode and train the model again for a few runs sequentially. This configuration is defined because switching back and forth between different classes more frequently to mimic shuffling the data is cumbersome in the real world. If a new class label is provided, TinyOL will automatically accommodate its layer structure for the new class. Otherwise, the system will update the weights at each run incrementally. 

We compare the model's performance under online and offline/batch setting. It is to mention that the underlying concept between online and offline training is quite different. Thus an intuitive result is hard to provide. When a model is trained incrementally, the data sample is discarded after the weights are adapted. That means each data will only be used to train the model once. For each data pairs, gradient descent will be called. Whereas in batch/offline mode, training data are reused for every epoch, and gradient descent will happen at each batch. We use F1-score and macro F1-score to evaluate the model on a pre-collected test dataset, where macro F1-score is the unweighted mean of F1-scores for all classes:

\begin{equation}
    F_1 = 2 \cdot  \frac{precision \cdot recall}{precision + recall}.
    \label{for:1}
\end{equation}

\begin{figure}[htbp]
  \centering%
    \begin{subfigure}{0.95\columnwidth}
    \includegraphics[width=1\linewidth]{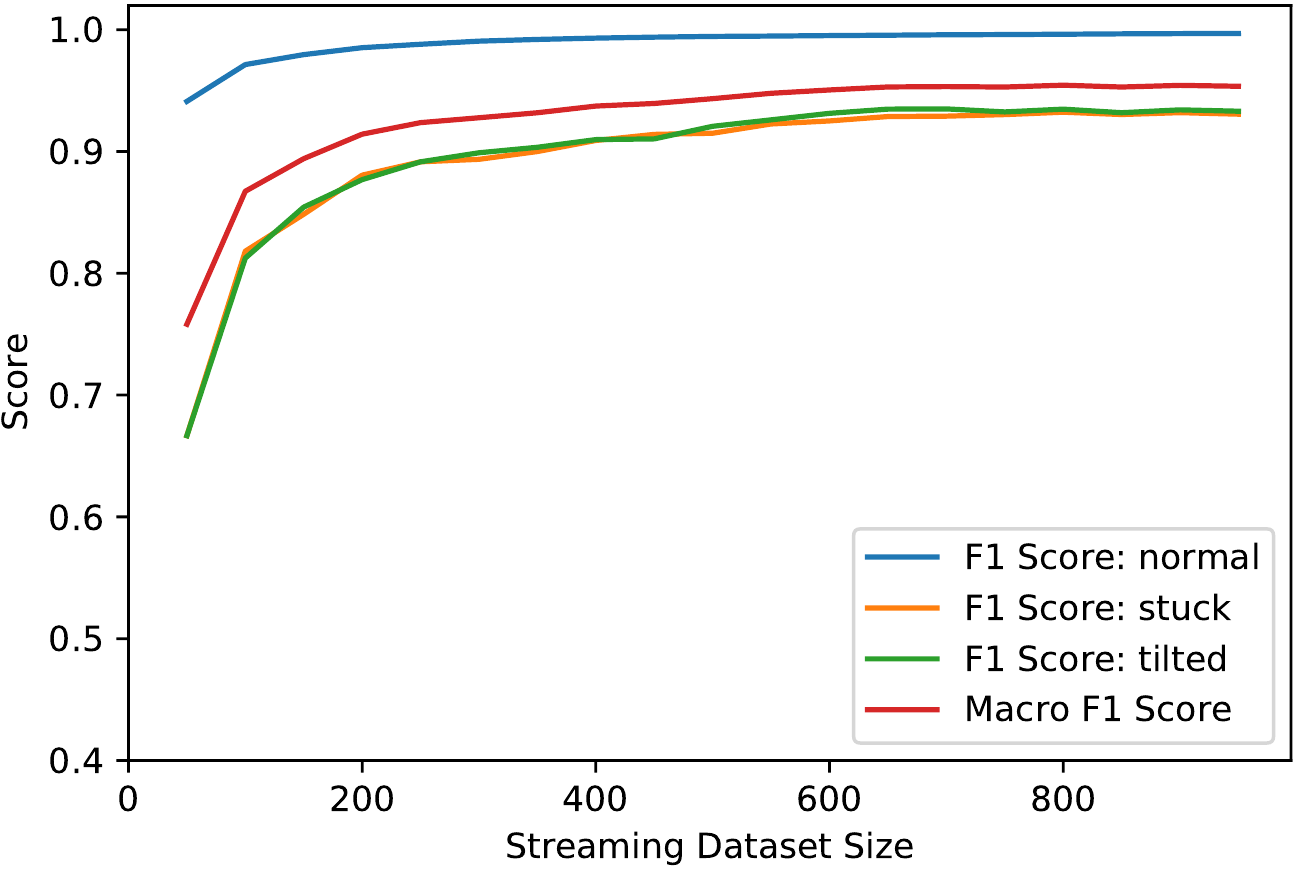} 
    \caption{Online learning: the F1-score of each classes and the averaged Macro F1-score. } 
    \end{subfigure} 
    \begin{subfigure}{0.95\columnwidth}
    \includegraphics[width=1\linewidth]{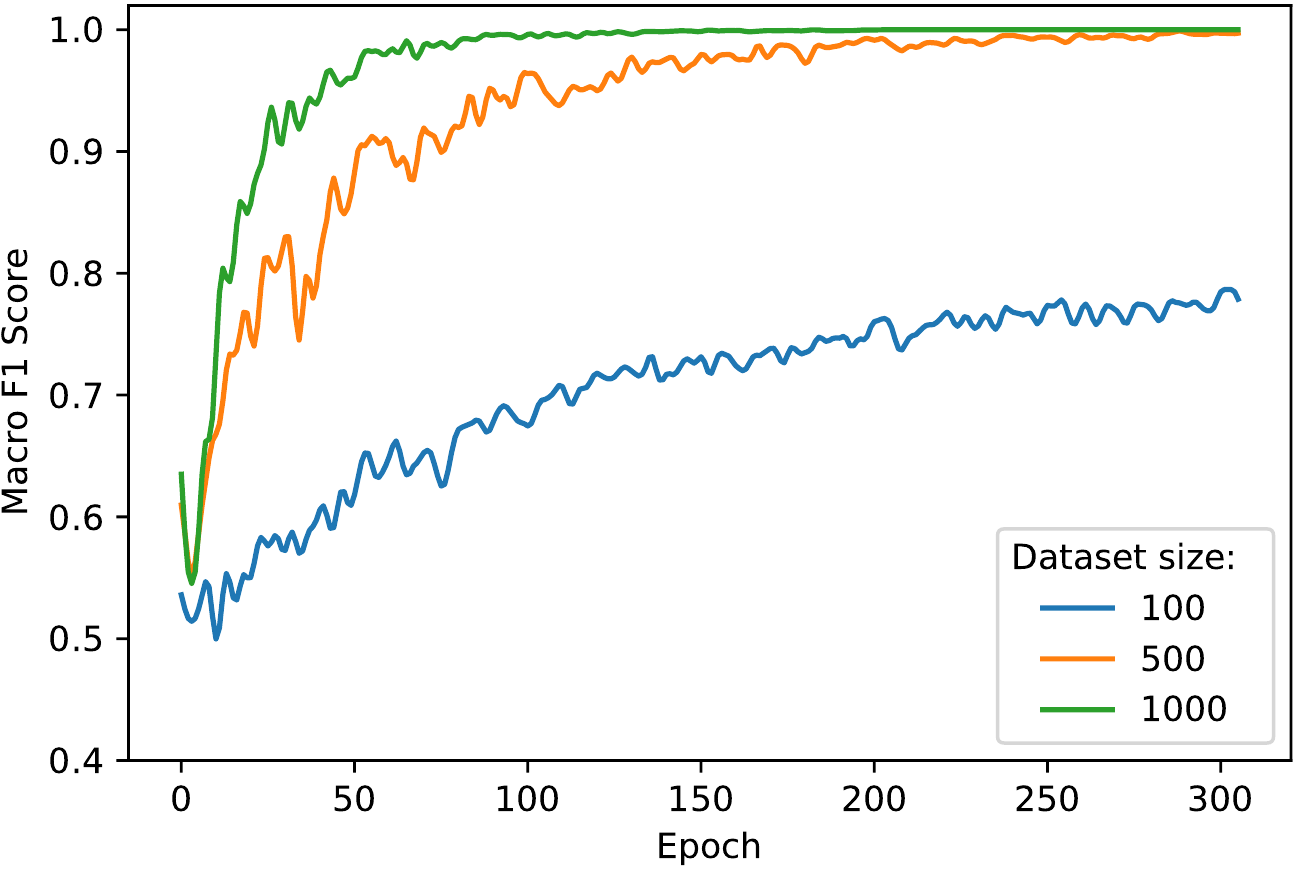} 
    \caption{Offline learning: the Macro F1-scores of the model trained on datasets of different size.} 
    \end{subfigure} 
    \caption{The performance comparison between online learning and offline learning.} 
    \label{f12}
\end{figure}
    
We evaluate the online model trained on streaming datasets of different sizes, that is, assessing the model after every 50 learning steps. As shown in the upper figure in Fig.~\ref{f12}, the online model performs better when it is trained on more streaming data. The F1-score on the normal class is always higher than the other two abnormal classes. This may be due to the autoencoder is trained only on the data from normal class, and the weights in the encoder are frozen during the post-training. Hence, the encoder cannot extract the features in the abnormal data efficiently. We might achieve a better performance if all the weights can be updated. Comparing the macro F1-score of online learning (upper) and offline learning (lower), we observe that offline batch learning outperforms online learning if the amount of data is enough and the model is trained for more than 50 epochs. However, performing batch training needs more computational resources since the model is trained on the whole dataset at each epoch. Considering the online model is trained incrementally with limited shuffling, the result is overall competitive. 



\section{Conclusions and Future Work}
\label{section:conclude}

The TinyML will play an essential role in the IoT by enabling large-scale ML applications.  The on-device training on MCUs, however, is still an open challenge. In this work, we propose a novel solution for on-device training by leveraging incremental online learning. We show that the accuracy of the model degrades when training and inference run on different datasets. By incorporating the proposed TinyOL system with an arbitrary existing NN, we can adapt the NN to new working conditions and improve its performance over time. Moreover, the system is featured with the flexibility to modify the layer structure on the fly. We proved the proposed solutions' effectiveness and feasibility on an Arduino board with a USB fan using autoencoder in two scenarios, i.e., anomaly detection and anomaly classification. We compared the performance between online learning and offline training. Although the model trained offline outperforms the model trained online when enough data and computational power are provided, TinyOL satisfies our expectation considering constrained resources of tiny devices.

Our future works include:
\begin{itemize}
\item Design efficient algorithms to empower full-size on-device training.
\item Support more optimizers and operations of neural networks, e.g., recurrent neural network.
\item Analyze the possibility to integrate online dimension reduction.
\item Explore the training with reduced numerical precision, e.g., on 8-bit MCUs.
\item Realize mass management of "intelligent" IoT devices.
\item Open source the clearance of the TinyOL repository.
\end{itemize}

\end{document}